\documentclass{article}
\usepackage{spconf,amsmath,graphicx}
\usepackage{booktabs} 
\usepackage{svg}

\title{Prior-enhanced Temporal Action Localization using Subject-aware Spatial Attention}

\name{Yifan Liu$^{\star}$ \qquad Youbao Tang$^{\dagger}$ \thanks{Corresponding authors: Youbao Tang \& Haoqian Wang} \qquad Ning Zhang$^{\dagger}$ \qquad Ruei-Sung Lin$^{\dagger}$ \qquad Haoqian Wang$^{\star}$}
  
  \address{$^{\star}$ Tsinghua University, Shenzhen, China \\
      $^{\dagger}$ PAII Inc. Palo Alto, CA, USA}

\begin{document}
%
\maketitle
\begin{abstract}
Temporal action localization (TAL) aims to detect the boundary and identify the class of each action instance in a long untrimmed video. Current approaches treat video frames homogeneously, and tend to give background and key objects excessive attention. This limits their sensitivity to localize action boundaries. To this end, we propose a prior-enhanced temporal action localization method (PETAL), which only takes in RGB input and incorporates action subjects as priors. This proposal leverages action subjects' information with a plug-and-play subject-aware spatial attention module (SA-SAM) to generate an aggregated and subject-prioritized representation. Experimental results on THUMOS-14 and ActivityNet-1.3 datasets demonstrate that the proposed PETAL achieves competitive performance using only RGB features, \textit{e.g.}, boosting mAP by 2.41\% or 0.25\% over the state-of-the-art approach that uses RGB features or with additional optical flow features on the THUMOS-14 dataset.
\end{abstract}
\section{Introduction}
\label{sec:intro} 

Long video understanding has become an important task due to the explosive increase in the amount of video content in recent years. As one of the fundamental video understanding tasks, temporal action localization (TAL) focuses on two parallel tasks of an untrimmed long video: action type identification and temporal boundary detection. Currently, as the generated video footage is getting longer and longer, detecting the start and the end timestamp of action instances is more crucial than identifying the action.

A classic TAL pipeline has two key steps. First, snippets (usually 16 frames) are processed using a sliding window over the whole untrimmed video. Subsequently, each snippet goes into a video encoder to extract a global feature. Second, the extracted features are temporally arranged into action detectors to detect action instances. Recent works \cite{lin2017single, long2019gaussian, yang2020revisiting, liu2020PBRNet, lin2021AFSD, liu2022empirical, tallfromer, zhang2022actionformer, shi2022react, dai2022ms} are devoted to detecting action instances using single-stage fashion. These works have achieved more promising performance with higher computational efficiency over the two-stage counterparts \cite{lin2018bsn, chao2018rethinking, lin2019bmn, xu2020g, chen2022dcan}. The existing single-stage TAL methods are divided into 3 categories: anchor-based, anchor-free, and query-based method. Lin et al. \cite{lin2017single} proposed the first single-stage method with anchor windows. Long et al. \cite{long2019gaussian} incorporated multi-scale Gaussian kernels to dynamically generate anchors instead of using predefined fixed scale anchors. Lin et al. \cite{lin2021AFSD} proposed the first purely anchor-free method focusing on salient boundaries of actions. Zhang et al. \cite{zhang2022actionformer}  designed a multi-scale transformer as an anchor-free TAL method. Liu et al. \cite{liu2022empirical} illustrated a query-based method without post-processing taking inspiration from DETR \cite{carion2020DETR}.

\begin{figure}[t]
\flushleft
\includegraphics[width=0.49\textwidth]{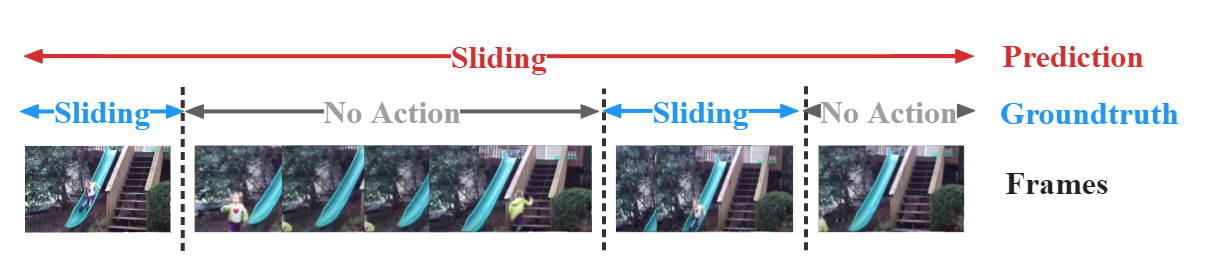}
\caption{An example that current approaches tend to give excessive attention to the background and key objects. Hence, No action is mistaken as sliding action as depicted.}
\label{fig:limitation}
\vspace{-0.3cm}
\end{figure}

One issue with the aforementioned approaches is the homogeneous frame process. Average pooling is utilized across the entire frame, and it fails in temporal boundary detection due to the uniform nature of the entire-frame processing without spatial priorities. As a result, sometimes background objects gain excessive attention and dominate the subject of the actual actions. This limits the existing TAL methods to be used in a more complex video setting. Fig.\ref{fig:limitation} shows an example of this issue, where the true action is a person \textbf{sliding} down the slide. Before or after the  \textbf{sliding} action, despite the presence of the slide object in the background, it should be considered as \textbf{no action}. However, the state-of-the-art (SOTA) approach \cite{zhang2022actionformer} classifies these \textbf{no action} segments as \textbf{sliding}.

Contrary to popular SOTA approaches of homogeneous frame process \cite{liu2022empirical, tallfromer, zhang2022actionformer, shi2022react, dai2022ms}, we believe that the TAL task needs a spatial prior to emphasize the action subjects. To achieve this, we propose a prior-enhanced temporal action localization method (PETAL) by incorporating a subject-aware spatial attention module (SA-SAM). Hence PETAL is able to explicitly consider the important action subjects that uniquely resemble the action itself. From our observation and analysis of the prominent TAL datasets \cite{caba2015activitynet, idrees2017thumos}, such an action subject is "human", and the actions of "human" should play a dominant role instead of being dominated by the background information. Therefore, the core motivation of our method is explicitly exploiting the dominant role of the action subjects for TAL.

The proposed PETAL utilizes a multi-person representation to explicitly leverage the dominant role of "human" and thus helps to aggregate the attended features of multiple people in space. We denote these attended features of individual people as \textbf{individual token}. The aggregated multi-person representation is named as \textbf{group token} in our method. Subsequently, the learned group token through SA-SAM is fed into the hierarchical temporal attention modules. This setting extracts the spatially abstracted information and allows the subsequent temporal attention to focus on the action subjects.

In summary, our contributions are as follows: 
\begin{itemize}

    \item We propose a prior-enhanced temporal action localization framework (PETAL). It focuses on action subjects in terms of a person and a group of people with spatially prioritized feature representation. Specifically, a simple subject-aware detection ranking module (SA-DRM) is introduced to emphasize the spatial locations occupied by the action subjects. Instead of a homogeneous process, an individual token is defined as the filtered feature of each subject after the ranking process. 
    \item A subject-aware spatial attention module (SA-SAM) is proposed to learn a multi-subject spatially prioritized group token from all individual tokens. Such a multi-head self-attention mechanism perceives the action collectively and emphasizes semantic priorities.
    \item Extensive experiments show that the proposed PETAL with RGB-only feature achieves better performance than the SOTA methods on THUMOS-14 and comparable results on ActivityNet-1.3.
    
\end{itemize}

\section{Method}
\label{sec:method}

\begin{figure}[t]
\flushleft
\centering
\includegraphics[width=0.48\textwidth]{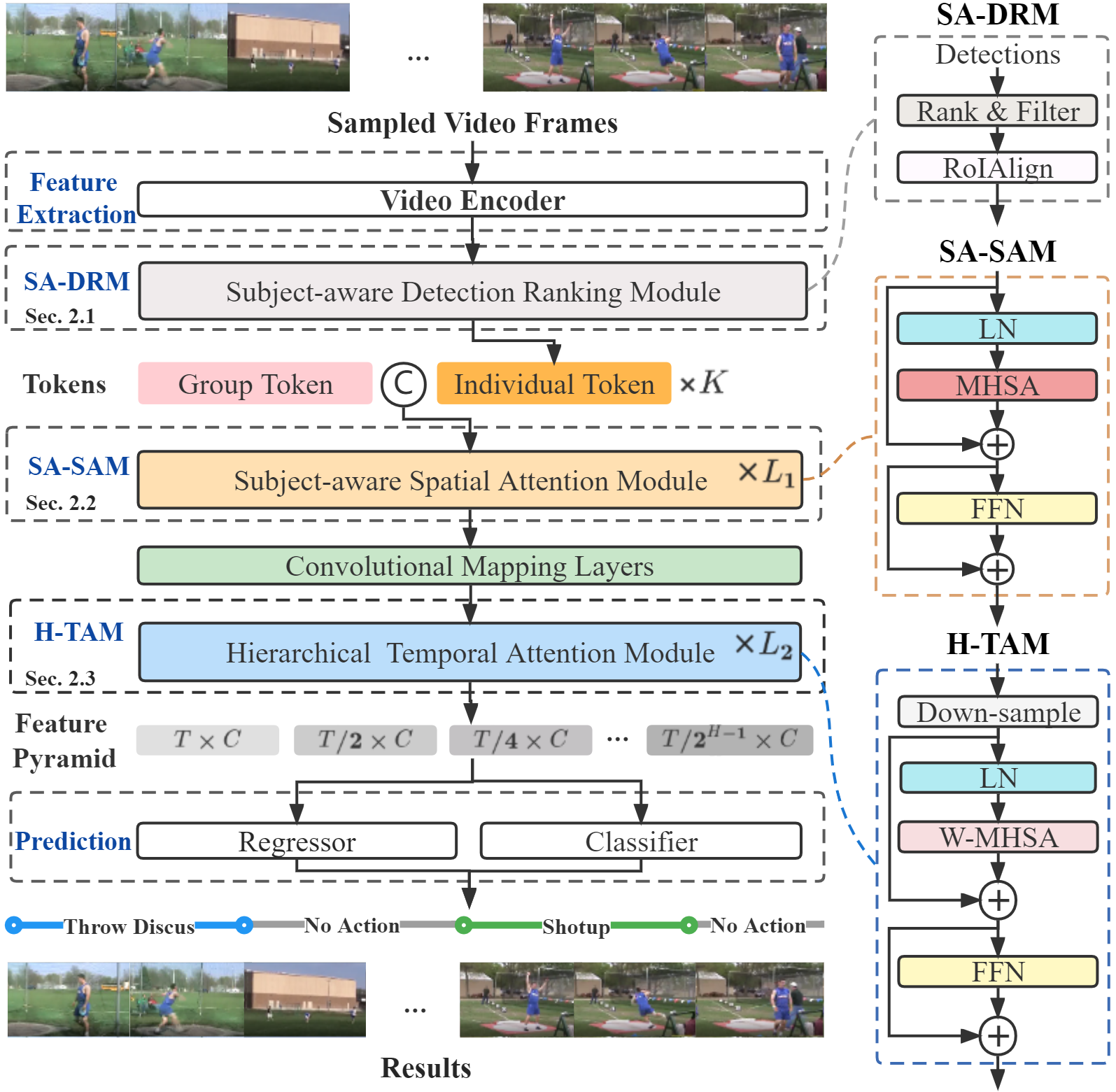}
\caption{Overview of the proposed PETAL.}
\label{fig:Overview}
\vspace{-8pt}
\end{figure}

In order to explicitly leverage the prior information of multiple action subjects information, we propose PETAL, whose overall pipeline is shown in Fig. \ref{fig:Overview}. Given an input video $X=\{x_1, x_2,\dots, x_{T_0}\}$, where $x_i \in \mathbf{R}^{H_0 \times W_0}$, $T_0$ indicates the frame number of the video, and $H_0 \times W_0$ is the spatial resolution, we first use a video encoder to extract snippet-level features $F=\{f_1, f_2, ..., f_T\}$, where $f_i \in \mathbf{R}^{H \times W \times D}$, $T$ indicates the number of snippets, $H \times W$ is the spatial resolution of the extracted features, and $D$ is the dimension of feature vectors. 

Once the snippet features are obtained, SA-DRM is used to filter all action subjects to $K$ most important subjects. This is explained in Sec. \ref{sec:ranking}. \textbf{Individual token} is introduced to represent the attended feature of each important subject generated from SA-DRM. Subsequently, the SA-SAM introduced in Sec. \ref{sec:SA} computes the  \textbf{group token}, a prior-enhanced feature containing multi-person information. In SA-SAM, we embed individual tokens into the group token using multi-head self-attention mechanism. After obtaining the final group token, we follow \cite{zhang2022actionformer} using two convolution layers for feature mapping and using hierarchical temporal attention module (H-TAM) to aggregate features across the temporal dimension. H-TAM is briefly introduced in Sec. \ref{sec:TA}. The stacked hierarchical temporal attention modules will generate a feature pyramid with descending temporal resolutions. With the feature pyramid, we use a regressor and a classifier containing 4 parameter-shared 1D convolution layers followed by a ReLU activation function for action temporal boundary regression and category classification, respectively.

\subsection{Subject-aware Detection Ranking Module (SA-DRM)}
\label{sec:ranking}

We observe that action subjects are likely to obtain high detection confidence scores and occupy large areas. SA-DRM is used to pick $K$ most important action subjects in each frame by ranking the detected bounding boxes based on the ratios of their sizes and the original image size. 

Subsequently, we use the RoIAlign \cite{he2017mask} operation to crop features from the corresponding snippet feature $f_i$ and use the spatial average pooling to get the attended feature $p_k^0, k=1, \dots, K$ for each ranked subject. This is the so-called individual token. If only $K_1 (< K)$ bounding boxes are detected in one frame, we mask $p_k^0, k=K_1+1, \dots, K$ to zero, and also mask their attention scores in subsequent SA-SAM.

\subsection{Subject-aware Spatial Attention Module (SA-SAM)}
\label{sec:SA}

SA-SAM aims at generating a group token $g$ to aggregate multiple action subject information from individual tokens with multi-head self-attention. The group token $g$ has the same dimension as each individual token. Multi-head self-attention (MHSA) is a feature aggregation mechanism that is designed to dynamically compute the weights for feature aggregation based on their similarities. This operation aggregates the subject-aware information contained in different individual tokens into a group token altogether. We formulate the self-attention mechanism as
\begin{equation}
\begin{aligned}
        Z &= \texttt{concat}(p_1, p_2, \dots, p_K, g) \\
        Q &= ZW_q, K= ZW_k, V=ZW_v \\
        O &= \texttt{softmax}(QK^T/\sqrt{D_q}) V
\end{aligned}
\end{equation}
where $W_q \in \mathbf{R}^{ D \times D_q}, W_k \in \mathbf{R}^{ D \times D_k}, W_v \in \mathbf{R}^{ D \times D_v}, D_q = D_k$ is the mapping parameters. Multi-head self-attention separates input tokens $Z \in \mathbf{R}^{(K+1) \times D}$ into several low embedding dimension features and sends them to different heads. Each head has its own mapping parameters. The output of multi-head self-attention can be obtained by concatenating the outputs from all heads.

With the formulated multi-head self-attention, SA-SAM is defined as follows. Assuming that $p_k^l, k=1, 2, \dots, K$ and $g^l$ denote the individual tokens and group token aggregated by the $l^{th}$ layer respectively, we formulate our stacked subject-aware spatial attention modules as
\begin{equation}
    \begin{aligned}
        Z^l &= \texttt{concat}(p_1^l, p_2^l, \dots, p_K^l, g^l) \\
        Z^{l+1} &= \texttt{MHSA}(\texttt{LN}(Z^l)) + Z^l \\
        Z^{l+1} &= \texttt{FFN}(\texttt{LN}(Z^{l+1})) + Z^{l+1} \\
        l&=0, 1, \dots, L_1-1
    \end{aligned}
\end{equation}
where $g^0$ is initialized by the global spatial average feature of the extracted video feature, $L_1$ is the number of layers, LN represents the layer normalization, and FFN means the feed forward network composed of linear layers.

\subsection{Hierarchical Temporal Attention Module (H-TAM)}
\label{sec:TA}

H-TAM expands the prior-enhanced features (\textit{i.e.}, the final learned group token $g^{L_1}$ of each snippet) into the temporal domain. We follow ActionFormer approach with windowing-based multi-head self-attention \cite{zhang2022actionformer}. Down-sampling is used to reduce the temporal resolution so that the fixed-length window attention can cover more time steps as the number of stacked layers of the temporal attention module increases. Fig. \ref{fig:Overview} illustrates this process in detail. Finally, We stack H-TAM for $L_2$ layers to generate a feature pyramid for regression and classification. When the down-sampling rate $\alpha$, i.e. stride of 1D convolution, equals 1, the H-TAM degenerates into a standard temporal attention module.

\subsection{Training And Inference}
\label{sec:training}
\textbf{Training Loss.} We use generalized IoU loss \cite{rezatofighi2019generalized} as regression loss and focal loss \cite{lin2017focal} as classification loss. We adopt the annotation  $(\texttt{class}, d^s_t, d^e_t)$ for all time steps $t=1,2, \dots, T$, where $d^s_t=t-\texttt{start}, d^e_t=\texttt{end}-t$. We only consider training loss when time step $t \in [\texttt{start}, \texttt{end}]$. The training loss is defined as:

\begin{equation}
    L = \sum_{l=1}^H{\sum_{t\in \mathcal{T}^+}{1 \over T^+}(L_{cls}^l + \lambda L_{reg}^l)}
\end{equation}
where $H$ indicates the height of the feature pyramid, and $\mathcal{T}^+ \subset \mathcal{T}=\{1,2, \dots, T\}$ denotes the set of time steps within any action instance. $\lambda$ is used to balance classification loss and regression loss.

\textbf{Inference.} Given output $(p, \hat{d}^s_t, \hat{d}^e_t)$ for time step $t$, we decode the output to predict segment $(p, t-\hat{d}^s_t, \hat{d}^e_t+t)$, where $p$ is the probabilities of all possible actions. Next, we perform Soft-NMS \cite{DBLP:conf/iccv/BodlaSCD17} to remove redundant results.

\section{Experiments}
\label{sec:exp}

\subsection{Datasets and Implementation Details}
\label{sec: exp-1}

\textbf{Datasets}
We conduct experiments on two commonly used datasets: THUMOS-14 dataset \cite{idrees2017thumos} and ActivityNet-1.3 dataset \cite{caba2015activitynet}. THUMOS-14 is an untrimmed video dataset that includes 220 and 212 videos in validation and testing sets from 20 classes. ActivityNet-1.3  contains 200 activity classes with 10,024 training and 4926 validation videos, respectively.

\textbf{Video Feature Extraction.}
We use the pre-extracted features for training. Specifically, we use VideoMAE \cite{tong2022videomae} to extract features for comparison. It only uses RGB input, which has 16x down-sampled spatial resolution. We use a sliding window of 16 frames with 75\% overlap to generate snippets for feature extraction on image resolution 224 $\times$ 224. We extract the VideoMAE feature at 15 fps. For pre-detected people bounding boxes, we use the center frame of each snippet as a keyframe to generate bounding boxes by using YOLOX \cite{yolox2021}.

\textbf{Network and Training Details.}
In our experiments, we build our network with $L_1$=8 stacked SA-SAMs, 2 convolution mapping layers, 2 stacked standard temporal attention modules, and 5 stacked H-TAMs with a down-sampling rate of 2 ($L_2$=7). The height of the feature pyramid is $H$=6. For training details, we set the initial learning rate to 0.0001 for THUMOS-14 and 0.001 for ActivityNet-1.3 with cosine annealing. We train our model using Adam optimizer \cite{DBLP:journals/corr/KingmaB14} for 35 epochs for THUMOS-14 and 15 epochs for ActivityNet-1.3 including 5 epochs' warm-up. We set the batch size to 2 for THUMOS-14 and 16 for ActivityNet-1.3, respectively. And we also use Exponential Moving Average (EMA) to stabilize training.

\textbf{Comparisons on THUMOS-14.}
The results of comparisons with SOTA methods on THUMOS-14 are shown in Table \ref{tab:sota-thumos14}. The proposed PETAL outperforms the previous SOTA results by 0.25\% of average mAP using only RGB features (no optical flow). We also implemented the  ActionFormer \cite{zhang2022actionformer} using the same RGB feature as ours, which is denoted * in Table\ref{tab:sota-thumos14}. This demonstrates that spatial-aware emphasis improves TAL performance.
\begin{table}[h]
\vspace{-10pt}
    \centering
    \caption{Comparison with State-of-the-Art on THUMOS-14. * refers to the method with our extracted feature. F denotes Flow. This and the following Table values are in percentile scale (\%)}
    \resizebox{\linewidth}{!}{
    \begin{tabular}{lccccccc}

    \toprule
        Method  &   Feature  &  0.3  & 0.4	& 0.5 & 0.6 & 0.7 & Avg \\
    \midrule
        GTAN \cite{long2019gaussian}& RGB & 57.8 &47.2 &38.8 & - & - & - \\
        A$^2$Net \cite{yang2020revisiting} & RGB+F &58.6 &54.1 &45.5 &32.5 &17.2& 41.6 \\
        PBRNet \cite{liu2020PBRNet} & RGB+F &58.5&54.6 &51.3 &41.8 &29.5 &47.1 \\
        AFSD \cite{lin2021AFSD}   & RGB+F & 67.3 &    62.4    &   55.5    &    43.7  &    31.1  &   52.0    \\
        E2ETAD \cite{liu2022empirical} & RGB & 69.4 &64.3& 56.0& 46.4& 34.9& 54.2 \\ 
        TALLFormer \cite{tallfromer} & RGB  & 76.0&-&63.2&- &34.5 &59.2 \\
        ActionFormer \cite{zhang2022actionformer}  & RGB+F     &  \underline{82.1} &\underline{77.8} &\textbf{71.0} &\textbf{59.4} &\textbf{43.9} &\underline{66.80} \\
        ActionFromer* \cite{zhang2022actionformer} & RGB &  81.00  & 76.82 &  68.66& 56.51 & 40.19 & 64.64\\

        PETAL (Ours)     & RGB &  \textbf{82.81}	&\textbf{78.90}&\underline{70.88}	&\underline{59.11}&\underline{43.57}	&\textbf{67.05}\\
    \bottomrule
    \end{tabular}}
    
    \label{tab:sota-thumos14}
\end{table}

\textbf{Comparisons on ActivityNet-1.3.}
The results of comparisons with SOTA methods on ActivityNet-1.3 are shown in Table \ref{tab:sota-anet}. Following previous works, the reported results use an ensemble of action classifiers \cite{zhao2017cuhk}. The results show that PETAL (35.34\%) achieves comparable results with SOTA ActionFormer using RGB and optical flow (35.6\%).  Without optical flow feature (denoted * in Table \ref{tab:sota-anet}), the proposed PETAL is 0.54\% higher than the ActionFormer approach. 

\begin{table}[h]
\vspace{-10pt}
    \centering
    \caption{Comparison with State-of-the-Art on ActivityNet-1.3. * refers to the method with our extracted feature. F denotes Flow. Avg mAP is calculated on mAP@[0.5:0.05:0.95].}
    \resizebox{\linewidth}{!}{
    \begin{tabular}{lccccc}

    \toprule
        Methods  &   Feature  &  0.5  & 0.75	& 0.95 & Avg \\
    \midrule
        GTAN \cite{long2019gaussian}     & RGB      & 52.61 &34.14 &8.91 &34.31 \\
        A$^2$Net \cite{yang2020revisiting} & RGB+F &   43.6& 28.7& 3.7& 27.8\\
        PBRNet \cite{liu2020PBRNet} & RGB+F & \textbf{53.96} &34.97& \underline{8.98}& 35.01 \\
        AFSD \cite{lin2021AFSD}   & RGB+F &  52.4& 35.3& 6.5 &34.4\\
        E2ETAD \cite{liu2022empirical} & RGB & 50.13& 35.78 &\textbf{10.52} &35.10 \\
        TALLFormer \cite{tallfromer} & RGB  &54.1 &\textbf{36.2} &7.9& \textbf{35.6} \\
        ActionFormer \cite{zhang2022actionformer}  & RGB+F  &53.5 &\textbf{36.2} &8.2 &\textbf{35.6} \\
        ActionFromer* \cite{zhang2022actionformer} & RGB & 53.33   & 35.54 &  6.76& 34.80 \\

        PETAL (Ours)     & RGB &  \underline{53.68}    &  \underline{36.15}    &   6.92  &   \underline{35.34}  \\
    \bottomrule
    \end{tabular}}
    \label{tab:sota-anet}
\end{table}

\textbf{Effectiveness on SA-DRM.} This ablation study focuses on Sec. \ref{sec:ranking}. In order to ignore irrelevant action subjects having too small occupations in each frame, we use SA-DRM to filter out $K$ most important action subjects. Assuming people that occupy more than 1/16 of one frame as positive people, statistical data on THUMOS-14 show that the maximum number of positive people in most videos is less than 8. Therefore, we walk the hyper-parameter $K$ in [3, 4, 5, 6, 7, 8] to investigate the effect of the number of important action subjects as shown in Table \ref{tab:ablation-2}. The results show that filtering 6 most important action subjects is most beneficial to subsequent SA-SAM with an average mAP of 66.99\%.

\begin{table}[h]
\vspace{-10pt}
    \centering
    \caption{Ablation study on the effect of the number of important action subjects in SA-DRM.}
    \resizebox{\linewidth}{!}{
    \begin{tabular}{lcccccc}
    \toprule
        Settings &  0.3  & 0.4	& 0.5 & 0.6 & 0.7 & Avg \\
\midrule
       $K=3$  &  \underline{82.70}	&78.14	&70.51	&58.26	&\underline{43.92} &66.71 \\
       $K=4$  &  \textbf{82.96}	&78.27	&70.15	&57.71	&43.34	&66.49\\
       $K=5$  &  82.19	&78.03	&\underline{70.88}	&57.89	&42.72	&66.34\\
       $K=6$  & 82.48	&\textbf{78.36}	&\textbf{70.99}	&\textbf{58.68}	&\textbf{44.46}	&\textbf{66.99}\\
       $K=7$  & 82.44	&\underline{78.32}	&70.65	&\underline{58.61}	&43.64	&\underline{66.73}\\
        $K=8$  & 82.20 &78.29	&70.59	&58.28	&42.91	&66.45\\
    \bottomrule
    \end{tabular}}

    \label{tab:ablation-2}
\end{table}

\textbf{Effectiveness on SA-SAM depth and embedding dimension of group token.} This ablation study focuses on Sec. \ref{sec:SA}. Fixing the number of important action subjects in SA-DRM, we also conduct extensive experiments to explore the effectiveness of the depth of SA-SAM and the group token dimension. Table \ref{tab:ablation-3} depicts the results with the following summary: (1) The original embedding dimension (768) of extracted feature achieves the best performance in SA-SAM. (2) Stacked SA-SAMs for 6 layers can achieve promising results of 66.99\% on average mAP with the best mAP of 44.46\% given tIOU=0.7. Stacking SA-SAM for 8 layers obtains the best average mAP of 67.05\%.

\begin{table}[h]
\vspace{-10pt}
    \centering
    \caption{Ablation study on the effect of depth of SA-SAM and embedding dimension of group token.}
    \resizebox{\linewidth}{!}{
    \begin{tabular}{llcccccc}
    \toprule
        Depth &  Dim &  0.3  & 0.4	& 0.5 & 0.6 & 0.7 & Avg \\
    \midrule
        4  &  768  &  81.97	&77.58	&70.12	&58.32	&43.13	& 66.23\\
        6  &  768  &  \underline{82.48}	&\underline{78.36}	&\textbf{70.99}	&\underline{58.68}	&\textbf{44.46}	& \underline{66.99}\\
        8  &  768  &  \textbf{82.81}	&\textbf{78.90}	&\underline{70.88}	&\textbf{59.11}	&\underline{43.57}	&\textbf{67.05}\\
        6  &  1024 &  81.96	&77.77	&69.62	&56.52	&42.52	&65.68\\
        6  &  512  &  81.70	&77.51	&69.22	&57.07	&42.96	&65.69\\
        6  &  256  &  81.41	&77.30	&68.92	&56.53	&41.58	&65.15\\
    \bottomrule
    \end{tabular}}
    
    \label{tab:ablation-3}
\end{table}

\vspace{-4pt}
\section{Conclusion}
\label{sec:conclusion}
In summary, we proposed a novel TAL framework PETAL. PETAL effectively introduces prior information of action subjects to avoid superfluous attention to the background and key objects. In addition, we also propose a plug-and-play SA-SAM to aggregate action subjects' information. Extensive experiments show PETAL achieves competitive results on two commonly used TAL datasets. Future work will consider incorporating optical flow features under the PETAL pipeline for a better perception of actions and performance.

\bibliographystyle{IEEEbib}
\bibliography{refs}

\begin{thebibliography}{10}

\bibitem{lin2017single}
Tianwei Lin, Xu~Zhao, and Zheng Shou,
\newblock ``Single shot temporal action detection,''
\newblock in {\em Proceedings of the ACM MM}, 2017, pp. 988--996.

\bibitem{long2019gaussian}
Fuchen Long, Ting Yao, Zhaofan Qiu, et~al.,
\newblock ``Gaussian temporal awareness networks for action localization,''
\newblock in {\em Proceedings of the IEEE CVPR}, 2019, pp. 344--353.

\bibitem{yang2020revisiting}
Le~Yang, Houwen Peng, Dingwen Zhang, et~al.,
\newblock ``Revisiting anchor mechanisms for temporal action localization,''
\newblock {\em IEEE TIP}, vol. 29, pp. 8535--8548, 2020.

\bibitem{liu2020PBRNet}
Qinying Liu and Zilei Wang,
\newblock ``Progressive boundary refinement network for temporal action
  detection,''
\newblock in {\em Proceedings of the AAAI}, 2020, vol.~34, pp. 11612--11619.

\bibitem{lin2021AFSD}
Chuming Lin, Chengming Xu, Donghao Luo, et~al.,
\newblock ``Learning salient boundary feature for anchor-free temporal action
  localization,''
\newblock in {\em Proceedings of the IEEE CVPR}, 2021, pp. 3320--3329.

\bibitem{liu2022empirical}
Xiaolong Liu, Song Bai, and Xiang Bai,
\newblock ``An empirical study of end-to-end temporal action detection,''
\newblock in {\em Proceedings of the IEEE CVPR}, 2022, pp. 20010--20019.

\bibitem{tallfromer}
Feng Cheng and Gedas Bertasius,
\newblock ``Tallformer: Temporal action localization with a long-memory
  transformer,''
\newblock in {\em Proceedings of the ECCV}, 2022, pp. 503--521.

\bibitem{zhang2022actionformer}
Chenlin Zhang, Jianxin Wu, and Yin Li,
\newblock ``Actionformer: Localizing moments of actions with transformers,''
\newblock in {\em Proceedings of the ECCV}, 2022.

\bibitem{shi2022react}
Dingfeng Shi, Yujie Zhong, Qiong Cao, et~al.,
\newblock ``React: Temporal action detection with relational queries,''
\newblock in {\em Proceedings of the ECCV}, 2022.

\bibitem{dai2022ms}
Rui Dai, Srijan Das, Kumara Kahatapitiya, et~al.,
\newblock ``Ms-tct: Multi-scale temporal convtransformer for action
  detection,''
\newblock in {\em Proceedings of the IEEE CVPR}, 2022, pp. 20041--20051.

\bibitem{lin2018bsn}
Tianwei Lin, Xu~Zhao, Haisheng Su, et~al.,
\newblock ``Bsn: Boundary sensitive network for temporal action proposal
  generation,''
\newblock in {\em Proceedings of the ECCV}, 2018, pp. 3--19.

\bibitem{chao2018rethinking}
Yu-Wei Chao, Sudheendra Vijayanarasimhan, Bryan Seybold, et~al.,
\newblock ``Rethinking the faster r-cnn architecture for temporal action
  localization,''
\newblock in {\em Proceedings of the IEEE CVPR}, 2018, pp. 1130--1139.

\bibitem{lin2019bmn}
Tianwei Lin, Xiao Liu, Xin Li, et~al.,
\newblock ``Bmn: Boundary-matching network for temporal action proposal
  generation,''
\newblock in {\em Proceedings of the IEEE ICCV}, 2019, pp. 3889--3898.

\bibitem{xu2020g}
Mengmeng Xu, Chen Zhao, David~S Rojas, et~al.,
\newblock ``G-tad: Sub-graph localization for temporal action detection,''
\newblock in {\em Proceedings of the IEEE CVPR}, 2020, pp. 10156--10165.

\bibitem{chen2022dcan}
Guo Chen, Yin-Dong Zheng, Limin Wang, et~al.,
\newblock ``Dcan: improving temporal action detection via dual context
  aggregation,''
\newblock in {\em Proceedings of the AAAI}, 2022, vol.~36, pp. 248--257.

\bibitem{carion2020DETR}
Nicolas Carion, Francisco Massa, Gabriel Synnaeve, et~al.,
\newblock ``End-to-end object detection with transformers,''
\newblock in {\em Proceedings of the ECCV}. Springer, 2020, pp. 213--229.

\bibitem{caba2015activitynet}
Bernard~Ghanem Fabian Caba~Heilbron, Victor~Escorcia et~al.,
\newblock ``Activitynet: A large-scale video benchmark for human activity
  understanding,''
\newblock in {\em Proceedings of the IEEE CVPR}, 2015, pp. 961--970.

\bibitem{idrees2017thumos}
Haroon Idrees, Amir~R Zamir, Yu-Gang Jiang, et~al.,
\newblock ``The thumos challenge on action recognition for videos “in the
  wild”,''
\newblock {\em Computer Vision and Image Understanding}, vol. 155, pp. 1--23,
  2017.

\bibitem{he2017mask}
Kaiming He, Georgia Gkioxari, Piotr Doll{\'a}r, et~al.,
\newblock ``Mask r-cnn,''
\newblock in {\em Proceedings of the IEEE ICCV}, 2017, pp. 2961--2969.

\bibitem{rezatofighi2019generalized}
Hamid Rezatofighi, Nathan Tsoi, JunYoung Gwak, et~al.,
\newblock ``Generalized intersection over union: A metric and a loss for
  bounding box regression,''
\newblock in {\em Proceedings of the IEEE CVPR}, 2019, pp. 658--666.

\bibitem{lin2017focal}
Tsung-Yi Lin, Priya Goyal, Ross Girshick, et~al.,
\newblock ``Focal loss for dense object detection,''
\newblock in {\em Proceedings of the IEEE ICCV}, 2017, pp. 2980--2988.

\bibitem{DBLP:conf/iccv/BodlaSCD17}
Navaneeth Bodla, Bharat Singh, Rama Chellappa, et~al.,
\newblock ``Soft-nms - improving object detection with one line of code,''
\newblock in {\em Proceedings of the IEEE ICCV}. 2017, pp. 5562--5570, {IEEE}
  Computer Society.

\bibitem{tong2022videomae}
Zhan Tong, Yibing Song, Jue Wang, et~al.,
\newblock ``Videomae: Masked autoencoders are data-efficient learners for
  self-supervised video pre-training,''
\newblock {\em arXiv preprint arXiv:2203.12602}, 2022.

\bibitem{yolox2021}
Zheng Ge, Songtao Liu, Feng Wang, et~al.,
\newblock ``Yolox: Exceeding yolo series in 2021,''
\newblock {\em arXiv preprint arXiv:2107.08430}, 2021.

\bibitem{DBLP:journals/corr/KingmaB14}
Diederik~P. Kingma and Jimmy Ba,
\newblock ``Adam: {A} method for stochastic optimization,''
\newblock in {\em Proceedings of the ICLR}, Yoshua Bengio and Yann LeCun, Eds.,
  2015.

\bibitem{zhao2017cuhk}
Yue Zhao, Bowen Zhang, Zhirong Wu, et~al.,
\newblock ``Cuhk \& ethz \& siat submission to activitynet challenge 2017,''
\newblock {\em arXiv preprint arXiv:1710.08011}, vol. 8, no. 8, 2017.

\end{thebibliography}

\end{document}